\documentclass[10pt,twocolumn,letterpaper]{article}
\pdfoutput=1

\usepackage{pifont}
\usepackage{xcolor}
\usepackage{makecell}
\usepackage{bm}
\usepackage{wacv}
\usepackage{times}
\usepackage{epsfig}
\usepackage{graphicx}
\usepackage{amsmath}
\usepackage{amssymb}
\usepackage{float}
\usepackage[subrefformat=parens,labelformat=parens]{subfig}

\usepackage{cellspace}
\usepackage{booktabs}       
\usepackage{threeparttable}
\usepackage{multirow}
\usepackage{paralist}
\usepackage[english]{babel}
\usepackage{marvosym}

\def\wacvPaperID{396} 

\wacvfinalcopy 

\ifwacvfinal
\def\assignedStartPage{1} 
\fi

\ifwacvfinal
\usepackage[breaklinks=true,bookmarks=false]{hyperref}
\else
\usepackage[pagebackref=true,breaklinks=true,colorlinks,bookmarks=false]{hyperref}
\fi

\usepackage{tablefootnote}
\usepackage{cleveref}  
\crefformat{table}{Table~#2#1#3}
\crefformat{section}{Section~#2#1#3}
\crefformat{figure}{Fig.~#2#1#3}
\crefformat{subfigure}{Fig.~#2#1#3}
\crefformat{equation}{Eq.~(#2#1#3)}
\crefformat{algorithm}{Algorithm~#2#1#3}
\crefformat{appendix}{Appendix #2#1#3}

\ifwacvfinal
\else
\fi

\begin{document}

\setlength{\abovedisplayskip}{.5\baselineskip}
\setlength{\belowdisplayskip}{.5\baselineskip}
\definecolor{myred}{rgb}{239, 0, 0}
\newcommand{\xmark}{\textcolor{myred}{\ding{54}}}

\title{Attentional Feature Fusion}


\author{Yimian Dai$^{1}$ \quad\quad Fabian Gieseke$^{2,3}$ \quad\quad Stefan Oehmcke$^3$\quad\quad Yiquan Wu$^{1}$ \quad\quad Kobus Barnard$^{4}$\\ [0.1in]
\normalsize
$^1$College of Electronic and Information Engineering, Nanjing University of Aeronautics and Astronautics, Nanjing, China \\
$^2$Department of Information Systems, University of M\"unster, M\"unster, Germany\\
$^3$Department of Computer Science, University of Copenhagen, Copenhagen, Denmark\\
$^4$Department of Computer Science, University of Arizona, Tucson, AZ, USA\\
}

\maketitle

\begin{abstract}
Feature fusion, the combination of features from different layers or branches, is an omnipresent part of modern network architectures. 
It is often implemented via simple operations, such as summation or concatenation, but this might not be the best choice.
In this work, we propose a uniform and general scheme, namely attentional feature fusion, which is applicable for most common scenarios, including feature fusion induced by short and long skip connections as well as within Inception layers.
To better fuse features of inconsistent semantics and scales, we propose a multi-scale channel attention module, which addresses issues that arise when fusing features given at different scales.
We also demonstrate that the initial integration of feature maps can become a bottleneck and that this issue can be alleviated by adding another level of attention, which we refer to as iterative attentional feature fusion.
With fewer layers or parameters, our models outperform state-of-the-art networks on both CIFAR-100 and ImageNet datasets, which suggests that more sophisticated attention mechanisms for feature fusion hold great potential to consistently yield better results compared to their direct counterparts.
Our codes and trained models are available online\footnote{https://github.com/YimianDai/open-aff}.
\vspace{-0.35cm}

\end{abstract}


\section{Introduction}\label{sec:intro}

Convolutional neural networks (CNNs) have seen a significant improvement of the representation power by going deeper \cite{CVPR16ResNetV1}, going wider \cite{CVPR15GoogLeNet,BMVC16WideResNet}, increasing cardinality \cite{CVPR17ResNeXt}, and refining features dynamically \cite{CVPR18SENet}, corresponding to advances in many computer vision tasks.

Apart from these strategies, in this paper, we investigate a different component of the network, \emph{feature fusion}, to further boost the representation power of CNNs.
Whether explicit or implicit, intentional or unintentional, feature fusion is omnipresent for modern network architectures and has been studied extensively in the previous literature \cite{CVPR15GoogLeNet,NIPS15Highway,CVPR16ResNetV1,MICCAI15UNet,CVPR17FPN}. 
For instance, in the InceptionNet family \cite{CVPR15GoogLeNet,CVPR16Inception,AAAI17InceptionV4}, the outputs of filters with multiple sizes on the same level are fused to handle the large variation of object size. 
In Residual Networks (ResNet)~\cite{CVPR16ResNetV1,ECCV16ResNetV2} and its follow-ups \cite{BMVC16WideResNet,CVPR17ResNeXt}, the identity mapping features and residual learning features are fused as the output via short skip connections, enabling the training of very deep networks. 
In Feature Pyramid Networks (FPN) \cite{CVPR17FPN} and U-Net \cite{MICCAI15UNet}, low-level features and high-level features are fused via long skip connections to obtain high-resolution and semantically strong features, which are vital for semantic segmentation and object detection.
However, despite its prevalence in modern networks, most works on feature fusion focus on constructing sophisticated \emph{pathways} to combine features in different kernels, groups, or layers.
The feature fusion \emph{method} has rarely been addressed and is usually implemented via simple operations such as addition or concatenation, which merely offer a fixed linear aggregation of feature maps and are entirely unaware of whether this combination is suitable for specific objects.

Recently, Selective Kernel Networks (SKNet) \cite{CVPR19SKNet} and ResNeSt \cite{arXiv20ResNeSt} have been proposed to render dynamic weighted averaging of features from multiple kernels or groups \emph{in the same layer} based on the \emph{global} channel attention mechanism \cite{CVPR18SENet}. 
Although such attention-based methods present nonlinear approaches for feature fusion, they still suffer from the following shortcomings:
\begin{compactenum}
  \item \emph{Limited scenarios:}
  SKNet and ResNeSt only focus on the soft feature selection in the same layer, whereas the \emph{cross-layer} fusion in skip connections has not been addressed, leaving their schemes quite heuristic. 
  Despite having different scenarios, all kinds of feature fusion implementations face the same challenge, in essence, that is, how to integrate features of different scales for better performance. 
  A module that can overcome the semantic inconsistency and effectively integrate features of different scales should be able to consistently improve the quality of fused features in various network scenarios.
  However, so far, there is still a lack of a generalized approach that can unify different feature fusion scenarios in a consistent manner.

  \item \emph{Unsophisticated initial integration:}
  To feed the received features into the attention module, SKNet introduces another phase of feature fusion in an involuntary but inevitable way, which we call \emph{initial integration} and is implemented by addition. 
  Therefore, besides the design of the attention module, as its input, the initial integration approach also has a large impact on the quality of fusion weights.
  Considering the features may have a large inconsistency on the scale and semantic level, an unsophisticated initial integration strategy ignoring this issue can be a bottleneck. 

  \item \emph{Biased context aggregation scale:}
  The fusion weights in SKNet and ResNeSt are generated via the global channel attention mechanism \cite{CVPR18SENet}, which is preferred for information that distributes more globally.
  However, objects in the image can have an extremely large variation in size. 
  Numerous studies have emphasized this issue that arises when designing CNNs, i.e., that the receptive fields of predictors should match the object scale range \cite{ICCV17S3FD,CVPR18SNIP,NIPS18SNIPER,ICCV19Trident}. 
  Therefore, merely aggregating contextual information on a global scale is too biased and weakens the features of small objects. 
  This gives rise to the question if a network can dynamically and adaptively fuse the received features in a contextual scale-aware way.
\end{compactenum}

Motivated by the above observations, we present the \emph{attentional feature fusion} (AFF) module, trying to answer the question of how a unified approach for all kinds of feature fusion scenarios should be and address the problems of contextual aggregation and initial integration. 
The AFF framework generalizes the attention-based feature fusion from the same-layer scenario to cross-layer scenarios including short and long skip connections, and even the initial integration inside AFF itself. 
It provides a universal and consistent way to improve the performance of various networks, e.g., InceptionNet, ResNet, ResNeXt \cite{CVPR17ResNeXt}, and FPN, by simply replacing existing feature fusion operators with the proposed AFF module.
Moreover, the AFF framework supports to gradually refine the initial integration, namely the input of the fusion weight generator, by iteratively integrating the received features with another AFF module, which we refer to as \emph{iterative attentional feature fusion} (iAFF).

To alleviate the problems arising from scale variation and small objects, we advocate the idea that attention modules should also aggregate contextual information from different receptive fields for objects of different scales. 
More specifically, we propose the \emph{Multi-Scale Channel Attention Module} (MS-CAM), a simple yet effective scheme to remedy the feature inconsistency across different scales for attentional feature fusion.
Our key observation is that scale is not an issue exclusive to the spatial attention, and the channel attention can also have scales other than the global by varying the spatial pooling size.
By aggregating the multi-scale context information along the channel dimension, MS-CAM can simultaneously emphasize large objects that distribute more globally and highlight small objects that distribute more locally, facilitating the network to recognize and detect objects under extreme scale variation.

\section{Related Work}
\subsection{Multi-scale Attention Mechanism}

The scale variation of objects is one of the key challenges in computer vision. 
To remedy this issue, an intuitive way is to leverage multi-scale image pyramids \cite{TPAMI17ImagePyramid,CVPR16AttentionScale}, in which objects are recognized at multiple scales and the predictions are combined using non-maximum suppression. 
The other line of effort aims to exploit the inherent multi-scale, hierarchical feature pyramid of CNNs to approximate image pyramids, in which features from multiple layers are fused to obtain semantic features with high resolutions \cite{CVPR15Hypercolumns,MICCAI15UNet,CVPR17FPN}.

The attention mechanism in deep learning, which mimics the human visual attention mechanism \cite{CVPR19DAVSOD,CVPR20JLDCF}, is originally developed on a global scale. 
For example, the matrix multiplication in self-attention draws global dependencies of each word in a sentence \cite{NIPS17Attention} or each pixel in an image \cite{CVPR19DualAttention,CVPR18NonLocalNet,ICCV19AugmentedConv}. 
The Squeeze-and-Excitation Networks (SENet) squeeze global spatial information into a channel descriptor to capture channel-wise dependencies \cite{CVPR18SENet}.
Recently, researchers start to take into account the scale issue of attention mechanisms.
Similar to the above-mentioned approaches handling scale variation in CNNs, multi-scale attention mechanisms are achieved by either feeding multi-scale features into an attention module or combining feature contexts of multiple scales inside an attention module.
In the first type, the features at multiple scales or their concatenated result are fed into the attention module to generate multi-scale attention maps, while the scale of feature context aggregation inside the attention module remains single \cite{CVPR16AttentionScale,CVPR17MultiContextAttention,TMI19DeepAttentiveFeatures,ITSC19MultiScaleAttention,Sinha2020Apr,tao2020hierarchical}. 
The second type, which is also referred to as multi-scale spatial attention, aggregates feature contexts by convolutional kernels of different sizes \cite{BMVC18PAN} or from a pyramid \cite{BMVC18PAN,CVPR19PyramidAttention} inside the attention module . 

The proposed MS-CAM follows the idea of ParseNet \cite{arXiv15ParseNet} with combining local and global features in CNNs and the idea of spatial attention with aggregating multi-scale feature contexts inside the attention module, but differ in at least two important aspects: 
1) MS-CAM puts forward the scale issue in channel attention and is achieved by point-wise convolution rather than kernels of different sizes.
2) instead of in the backbone network, MS-CAM aggregates local and global feature contexts inside the channel attention module.
To the best of our knowledge, the multi-scale channel attention has never been discussed before.

\subsection{Skip Connections in Deep Learning}

Skip connection has been an essential component in modern convolutional networks.
Short skip connections, namely the identity mapping shortcuts added inside Residual blocks, provide an alternative path for the gradient to flow without interruption during backpropagation \cite{CVPR16ResNetV1,CVPR17ResNeXt,BMVC16WideResNet}.
Long skip connections help the network to obtain semantic features with high resolutions by bridging features of finer details from lower layers and high-level semantic features of coarse resolutions \cite{CVPR17DenseNet,CVPR17FPN,MICCAI15UNet,CVPR15FCN}.
Despite being used to combine features in various pathways \cite{Neurocomputing19Deepside}, the fusion of connected features is usually implemented via addition or concatenation, which allocate the features with fixed weights regardless of the variance of contents.
Recently, a few attention-based methods, e.g., Global Attention Upsample (GAU) \cite{BMVC18PAN} and Skip Attention (SA) \cite{SPL19SkipAttention}, have been proposed to use high-level features as guidance to modulate the low-level features in long skip connections. 
However, the fusion weights for the modulated features are still fixed.

To the best of our knowledge, it is the Highway Networks that first introduced a selection mechanism in short skip connections \cite{NIPS15Highway}.
To some extent, the \emph{attentional skip connections} proposed in this paper can be viewed as its follow-up, but differs in the three points:
1) Highway Networks employ a simple fully connected layer that can only generate a scalar fusion weight, while our proposed MS-CAM generates fusion weights as the same size of feature maps, enabling dynamic soft selections in an element-wise way. 
2) Highway Networks only use one input feature to generate weight, while our AFF module is aware of both features. 
3) We point out the importance of initial feature integration and the iAFF module is proposed as a solution.


\section{Multi-scale Channel Attention}

\subsection{Revisiting Channel Attention in SENet}

Given an intermediate feature $\mathbf{X} \in \mathbb{R}^{C \times H \times W}$ with $C$ channels and feature maps of size $H \times W$, the channel attention weights $\mathbf{w} \in \mathbb{R}^{C}$ in SENet can be computed as 
\begin{equation}
\mathbf{w} = \sigma  \left( \mathbf{g}(\mathbf{X}) \right) = \sigma \left( \mathcal{B} \left(\mathbf{W}_{2} \delta\left(\mathcal{B} \left(\mathbf{W}_{1} (g(\mathbf{X}))\right)\right)\right) \right),
\end{equation}
where $\mathbf{g}(\mathbf{X}) \in \mathbb{R}^{C}$ denotes the global feature context and
$g(\mathbf{X}) = \frac{1}{H \times W} \sum_{i = 1}^{H} \sum_{j = 1}^{W} \mathbf{X}_{[:, i, j]}$ is the global average pooling (GAP).
$\delta$ denotes the Rectified Linear Unit (ReLU) \cite{ICML10ReLU}, and $\mathcal{B}$ denotes the Batch Normalization (BN) \cite{ICML15BN}. 
$\sigma$ is the Sigmoid function.
This is achieved by a bottleneck with two fully connected (FC) layers,
where $\mathbf{W}_1 \in \mathbb{R}^{\frac{C}{r} \times C}$ is a dimension reduction layer, and $\mathbf{W}_2 \in \mathbb{R}^{C \times \frac{C}{r}}$ is a dimension increasing layer. $r$ is the channel reduction ratio.

We can see that the channel attention squeezes each feature map of size $H \times W$ into a scalar.
This extreme coarse descriptor prefers to emphasize large objects that distribute globally and can potentially wipe out most of the image signal present in a small object.
However, detecting very small objects stands out as the key performance bottleneck of state-of-the-art networks \cite{NIPS18SNIPER}. 
For example, the difficulty of COCO is largely due to the fact that most object instances are smaller than 1\% of the image area  \cite{ECCV14COCO,CVPR18SNIP}.
Therefore, global channel attention might not be the best choice. 
Multi-scale feature contexts should be aggregated inside the attention module to alleviate the problems arising from scale variation and small object instances. 

\subsection{Aggregating Local and Global Contexts}

In this part, we depict the proposed multi-scale channel attention module (MS-CAM) in detail. 
The key idea is that the channel attention can be implemented in multiple scales by varying the spatial pooling size. 
To maintain it as lightweight as possible, we merely add the local context to the global context inside the attention module. 
We choose the point-wise convolution (PWConv) as the local channel context aggregator, which only exploits point-wise channel interactions for each spatial position. 
To save parameters, the local channel context $\mathbf{L}(\mathbf{X}) \in \mathbb{R}^{C \times H \times W}$ is computed via a bottleneck structure as follows:
\begin{equation}
\mathbf{L}(\mathbf{X}) = \mathcal{B} \left(\mathrm{PWConv}_2 \left(\delta\left( \mathcal{B} \left(\mathrm{PWConv}_1 (\mathbf{X}) \right)\right)\right)\right)
\end{equation}
The kernel sizes of $\mathrm{PWConv}_1$ and $\mathrm{PWConv}_2$ are $\frac{C}{r} \times C \times 1 \times 1$ and $\mathrm{PWConv}_2$ is $C \times \frac{C}{r} \times 1 \times 1$, respectively.
It is noteworthy that $\mathbf{L}(\mathbf{X})$ has the same shape as the input feature, which can preserve and highlight the subtle details in the low-level features.
Given the global channel context $\mathbf{g}(\mathbf{X})$ and local channel context $\mathbf{L}(\mathbf{X})$, the refined feature $\mathbf{X}^{\prime} \in \mathbb{R}^{C \times H \times W}$ by MS-CAM can be obtained as follows:
\begin{equation}
\mathbf{X}^{\prime} = \mathbf{X} \otimes \mathbf{M}(\mathbf{X}) = \mathbf{X} \otimes \sigma \left(\mathbf{L}(\mathbf{X}) \oplus \mathbf{g}(\mathbf{X}) \right),
\end{equation}
where $\mathbf{M}(\mathbf{X}) \in \mathbb{R}^{C \times H \times W}$ denotes the attentional weights generated by MS-CAM.
$\oplus$ denotes the broadcasting addition and $\otimes$ denotes the element-wise multiplication.
\vspace*{-.5\baselineskip}

\begin{figure}[htbp]
  \centering
  \captionsetup[subfloat]{farskip=0pt,captionskip=0pt}
  \includegraphics[height=0.3\textwidth]{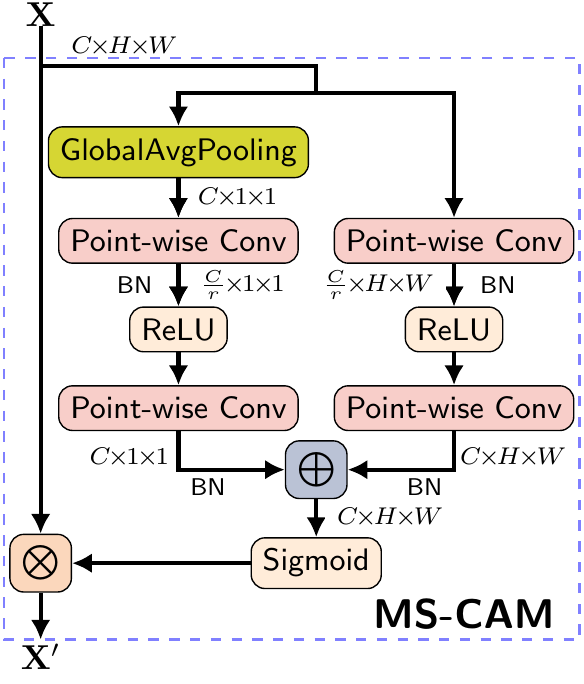}\\[-.5\baselineskip]
  \setlength{\belowcaptionskip}{-12pt}
  \caption{Illustration of the proposed MS-CAM}
  \label{fig:mscam}
\end{figure}

\section{Attentional Feature Fusion}

\subsection{Unification of Feature Fusion Scenarios}

\begin{figure}[htbp]
  \centering
  \captionsetup[subfloat]{farskip=0pt,captionskip=0pt}
  \subfloat[AFF]{
    \includegraphics[width=0.2\textwidth]{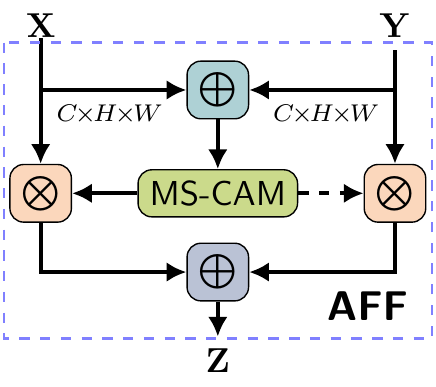}
    \label{subfig:aff}
  }
  \subfloat[iAFF]{
    \includegraphics[width=0.22\textwidth]{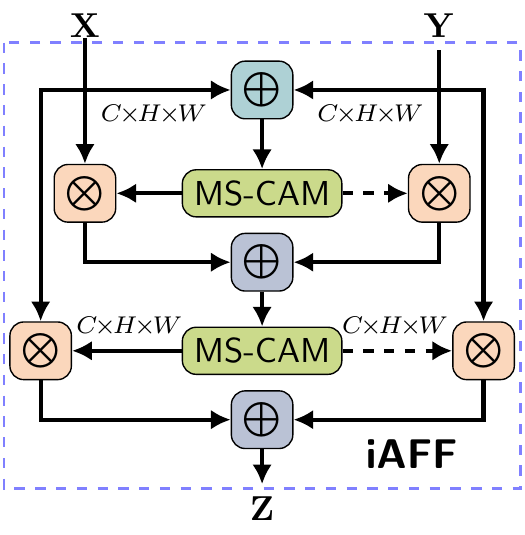}
    \label{subfig:iaff}
  }\\[-.5\baselineskip]  
  \setlength{\belowcaptionskip}{-2pt}  
  \caption{Illustration of the proposed AFF and iAFF}
  \label{fig:aff}  
  \vspace*{-.5\baselineskip}
\end{figure}

Given two feature maps $\mathbf{X}, \mathbf{Y} \in \mathbb{R}^{C \times H \times W}$, by default, we assume $\mathbf{Y}$ is the feature map with a larger receptive field. More specifically, 
\begin{compactenum}
  \item \emph{same-layer scenario}: $\mathbf{X}$ is the output of a $3\times 3$ kernel and $\mathbf{Y}$ is the output of a $5 \times 5$ kernel in InceptionNet; 
  \item \emph{short skip connection scenario}: $\mathbf{X}$ is the identity mapping, and $\mathbf{Y}$ is the learned residual in a ResNet block;
  \item \emph{long skip connection scenario}: $\mathbf{X}$ is the low-level feature map, and $\mathbf{Y}$ is the high-level semantic feature map in a feature pyramid.
\end{compactenum}
Based on the multi-scale channel attention module $\mathbf{M}$, \emph{Attentional Feature Fusion} (AFF) can be expressed as 
\begin{equation}
\mathbf{Z} = \mathbf{M}(\mathbf{X} \uplus \mathbf{Y}) \otimes \mathbf{X} + \left( 1 - \mathbf{M}(\mathbf{X} \uplus \mathbf{Y}) \right) \otimes \mathbf{Y},
\label{eq:aff}
\end{equation}
where $\mathbf{Z} \in \mathbb{R}^{C \times H \times W}$ is the fused feature, and $\uplus$ denotes the initial feature integration. In this subsection, for the sake of simplicity, we choose the element-wise summation as initial integration.
The AFF is illustrated in Fig.~\subref*{subfig:aff}, where the dashed line denotes $\mathbf{1} - \mathbf{M}(\mathbf{X} \uplus \mathbf{Y})$. It should be noted that the fusion weights $\mathbf{M}(\mathbf{X} \uplus \mathbf{Y})$ consists of real numbers between 0 and 1, so are the $1 - \mathbf{M}(\mathbf{X} \uplus \mathbf{Y})$, which enable the network to conduct a soft selection or weighted averaging between $\mathbf{X}$ and $\mathbf{Y}$.
\vspace*{-1.\baselineskip}

We summarize different formulations of feature fusion in deep networks in \cref{Tab:RelatedSkip}. 
$\mathbf{G}$ denotes the global attention mechanism. 
Although there are many implementation differences among multiple approaches for various feature fusion scenarios, once being abstracted into mathematical forms, these differences in details disappear. 
Therefore, it is possible to \emph{unify these feature fusion scenarios with a carefully designed approach}, thereby improving the performance of all networks by replacing original fusion operations with this unified approach.


\newcommand{\AddRows}{2}
\newcommand{\ConcatRows}{2}
\newcommand{\LinearRows}{\fpeval{\AddRows + \ConcatRows}}
\newcommand{\RefineXRows}{1}
\newcommand{\RefineYRows}{1}
\newcommand{\RefineXYRows}{1}
\newcommand{\RefineYXRows}{1}
\newcommand{\RefineRows}{1}
\newcommand{\SelectIncepRows}{3}
\newcommand{\SelectSkipRows}{2}
\newcommand{\SelectHighwayRows}{1}
\newcommand{\SKRows}{\fpeval{\SelectIncepRows}}
\newcommand{\ASKCRows}{0}
\newcommand{\SelectRows}{\fpeval{\SKRows + \SelectHighwayRows + \ASKCRows}}
\newcommand{\NonlinearRows}{\fpeval{\RefineRows + \SelectRows}}

\setlength{\tabcolsep}{2pt}

\begin{table*}[ht]
\centering
\caption{A brief overview of different feature fusion strategies in deep networks.\\[-.5\baselineskip]}
\label{Tab:RelatedSkip}
\small
\begin{tabular}{Sc Sc Sc Sl Sc} 
\toprule  
Context-aware                            & Type                                   & Formulation                               & Scenario               \& Reference & Example\\
\midrule 

\multirow{2}{*}{None} & \multirow{1}{*}{Addition}         & \multirow{1}{*}{$\mathbf{X} + \mathbf{Y}$} & Short Skip \cite{CVPR16ResNetV1,ECCV16ResNetV2}, Long Skip \cite{CVPR15FCN,CVPR17FPN}  & ResNet, FPN \\
                                  & \multirow{1}{*}{Concatenation} &
\multirow{1}{*}{$\mathbf{W}_{\mathbf{A}} \mathbf{X}_{:,i,j} + \mathbf{W}_{\mathbf{B}} \mathbf{Y}_{:,i,j}$}    
                                  & Same Layer \cite{CVPR15GoogLeNet}, Long Skip \cite{MICCAI15UNet,CVPR17DenseNet} & InceptionNet, U-Net \\
\midrule 
\multirow{3}{*}{Partially} & \multirow{1}{*}{Refinement} & $\mathbf{X} + \mathbf{G}(\mathbf{Y}) \otimes \mathbf{Y}$
                                        & Short Skip    \cite{CVPR18SENet,NIPS18GENet,ECCV18CBAM,BMVC18BAM} & SENet \\
                                        & \multirow{1}{*}{Modulation} &
 $\mathbf{G}(\mathbf{Y}) \otimes \mathbf{X} + \mathbf{Y}$
                                        & Long Skip                 \cite{BMVC18PAN} & GAU \\
                                        & \multirow{1}{*}{Soft Selection}           &
\multirow{1}{*}{$\mathbf{G}(\mathbf{X}) \otimes \mathbf{X} + (\mathbf{1} - \mathbf{G}(\mathbf{X})) \otimes \mathbf{Y}$}                 
                                        & Short Skip                         \cite{NIPS15Highway} & Highway Networks \\
\midrule 
\multirow{3}{*}{Fully} & \multirow{1}{*}{Modulation} & $\mathbf{G}(\mathbf{X}, \mathbf{Y}) \otimes \mathbf{X} + \mathbf{Y}$
                                        & Long Skip \cite{SPL19SkipAttention} & SA \\
                      & \multirow{2}{*}{Soft Selection} & $\mathbf{G}(\mathbf{X}+\mathbf{Y}) \otimes \mathbf{X} + (1 - \mathbf{G}(\mathbf{X}+\mathbf{Y})) \otimes \mathbf{Y}$
                                        & Same Layer    \cite{CVPR19SKNet,arXiv20ResNeSt} & SKNet \\
                      &                            & $\mathbf{M}(\mathbf{X} \uplus \mathbf{Y}) \otimes \mathbf{X} + (1 - \mathbf{M}(\mathbf{X} \uplus \mathbf{Y})) \otimes \mathbf{Y}$
                                        & Same Layer, Short Skip, Long Skip & \textbf{\textit{ours}}\\  
\bottomrule
\end{tabular}
\vspace*{-1.25\baselineskip}
\end{table*}

From \cref{Tab:RelatedSkip}, it can be further seen that apart from the implementation of the weight generation module $\mathbf{G}$, the state-of-the-art fusion schemes mainly differ in two crucial points:
(a)~the context-awareness level. 
Linear approaches like addition and concatenation are entirely contextual unaware. 
Feature refinement and modulation are non-linear, but only partially aware of the input feature maps. 
In most cases, they only exploit the high-level feature map.
Fully context-aware approaches utilize both input feature maps for guidance at the cost of raising the initial integration issue. 
(b)~Refinement vs modulation vs selection. 
The sum of weights applied to two feature maps in soft selection approaches are bound to 1, while this is not the case for refinement and modulation.

\subsection{Iterative Attentional Feature Fusion}

Unlike partially context-aware approaches \cite{BMVC18PAN}, fully context-aware methods have an inevitable issue, namely how to initially integrate input features. 
As the input of the attention module, the initial integration quality may profoundly affect final fusion weights. 
Since it is still a feature fusion problem, an intuitive way is to have another attention module to fuse input features. 
We call this two-stage approach \emph{iterative Attentional Feature Fusion} (iAFF), which is illustrated in Fig.~\subref*{subfig:iaff}. 
Then, the initial integration $\mathbf{X}\uplus\mathbf{Y}$ in \cref{eq:aff} can be reformulated as 
\begin{equation}
    \mathbf{X}\uplus\mathbf{Y} = \mathbf{M}(\mathbf{X}+\mathbf{Y}) \otimes \mathbf{X} + (1 - \mathbf{M}(\mathbf{X}+\mathbf{Y})) \otimes \mathbf{Y}    
\end{equation}

\subsection{Examples: InceptionNet, ResNet, and FPN}

\begin{figure}[tbp]
  \centering
  \captionsetup[subfloat]{farskip=0pt,captionskip=6pt}  
  \subfloat[AFF-Inception module]{
    \includegraphics[height=0.125\textwidth]{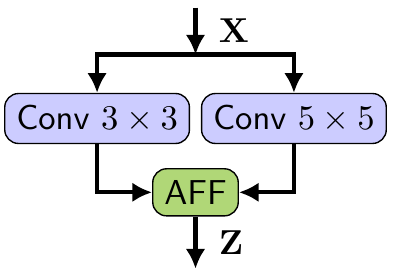}
  }\hfil
  \subfloat[AFF-ResBlock]{
    \includegraphics[height=0.125\textwidth]{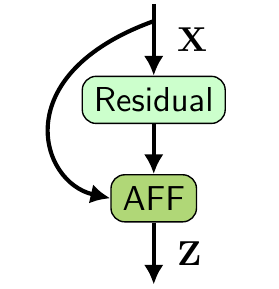}
  }

  \subfloat[AFF-FPN]{
    \includegraphics[height=0.1\textwidth]{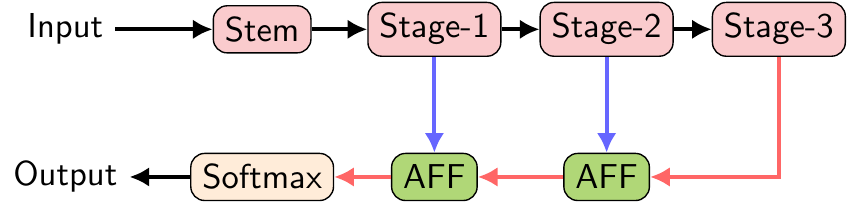}
  }  
  \vspace*{-.75\baselineskip}
  \setlength{\belowcaptionskip}{-1.25\baselineskip}  
  \caption{The schema of the proposed AFF-Inception module, AFF-ResBlock, and AFF-FPN. 
  The blue and red lines denote channel expansion and upsampling, respectively.
  }
  \label{fig:exemplars}
\end{figure}

To validate the proposed AFF/iAFF as a uniform and general scheme, we choose ResNet, FPN, and InceptionNet as examples for the most common scenarios: short and long skip connections as well as the same layer fusion.
It is straightforward to apply AFF/iAFF to existing networks by replacing the original addition or concatenation.
Specifically, we replace the concatenation in the InceptionNet module as well as the addition in ResNet block (ResBlock) and FPN to obtain the attentional networks, which we call AFF-Inception module, AFF-ResBlock, and AFF-FPN, respectively. 
This replacement and the schemes of our proposed architectures are shown in \cref{fig:exemplars}.
The iAFF is a particular case of AFF, so it does not need another illustration.

\section{Experiments}\label{sec:exp}




For experimental evaluation, we resort to the following benchmark datasets: CIFAR-100~\cite{Krizhevsky09CIFAR} and ImageNet~\cite{IJCV15ImageNet} for image classification in the same-layer InceptionNet and short-skip connection ResNet scenarios as well as StopSign (a subset of COCO dataset \cite{ECCV14COCO}) for semantic segmentation in the long-skip connection FPN scenario.
The detailed settings are listed in \cref{tab:setting}.
$b$ is the ResBlock number in each stage used to scale the network by depth.
Note that our CIFAR-100 experiments classify images into 20 super-classes, not 100 classes. It is a default setting of the CIFAR100 class in MXNet/Gluon. 
We didn't notice it until a bug issue in our github repo at the camera ready day.
However, since all the CIFAR-100 experiments are conducted on the same class number, our conclusion drawn from the experiment results still hold.
For more implementation details, please see the supplementary material and our code.

\setlength{\tabcolsep}{3pt}
\begin{table*}[htbp]
\centering
\caption{Experimental settings for the networks integrated with the proposed AFF/iAFF.
\\[-.5\baselineskip]}
\label{tab:setting}
\small
\begin{tabular}{Sc Sc Sc Sc Sc Sc Sc Sc Sc Sc Sc} 
\toprule  
\multirow{2}{*}{\makecell{Task}} & \multirow{2}{*}{\makecell{Dataset}} & \multirow{2}{*}{\makecell{Host Network}} & 
\multirow{2}{*}{\makecell{Fusing\\Scenario}} & 
\multirow{2}{*}{\makecell{$r$}} & \multirow{2}{*}{\makecell{Epochs}} & \multirow{2}{*}{\makecell{Batch\\Size}} & \multirow{2}{*}{\makecell{Optimizer}} & \multirow{2}{*}{\makecell{Learning\\Rate}} & \multirow{2}{*}{\makecell{Learning\\Rate Mode}} & \multirow{2}{*}{\makecell{Initialization}} \\
& & & & & & & & & & \\
\midrule 
\multirow{4}{*}{\makecell{Image\\Classification}}
& \multirow{3}{*}{CIFAR-100} & Inception-ResNet-20-$b$ & Same Layer & 4 & 400 & 128 & Nesterov & 0.2 & Step, $\gamma=0.1$ & Kaiming \\
& & ResNet-20-$b$ & Short Skip  & 4 & 400 & 128 & Nesterov & 0.2 & Step, $\gamma=0.1$ & Kaiming \\
& & ResNeXt-38-32x4d & Short Skip  & 16 & 400  & 128 & Nesterov & 0.2 & Step, $\gamma=0.1$ & Xavier \\
\cmidrule{2-11}
& ImageNet & ResNet-50 & Short Skip  & 16 & 160 & 128 & Nesterov & 0.075 & Cosine & Kaiming \\
\midrule 
\multirow{2}{*}{\makecell{Semantic\\Segmentation}} & \multirow{2}{*}{\makecell{StopSign}} & \multirow{2}{*}{\makecell{ResNet-20-$b$ + FPN}} & \multirow{2}{*}{\makecell{Long Skip}}  & \multirow{2}{*}{\makecell{4}} & \multirow{2}{*}{\makecell{300}} & \multirow{2}{*}{\makecell{32}} & \multirow{2}{*}{\makecell{AdaGrad}} & \multirow{2}{*}{\makecell{0.01}}& \multirow{2}{*}{\makecell{Poly}}  & \multirow{2}{*}{\makecell{Kaiming}} \\
& & & & & & & & & \\
\bottomrule
\end{tabular}
\vspace*{-.5\baselineskip}
\end{table*}

\subsection{Ablation Study}

\subsubsection{Impact of Multi-Scale Context Aggregation}

\begin{figure}[tbp]
  \centering
  \captionsetup[subfloat]{farskip=0pt,captionskip=0pt}
  \subfloat{
    \includegraphics[width=0.245\textwidth]{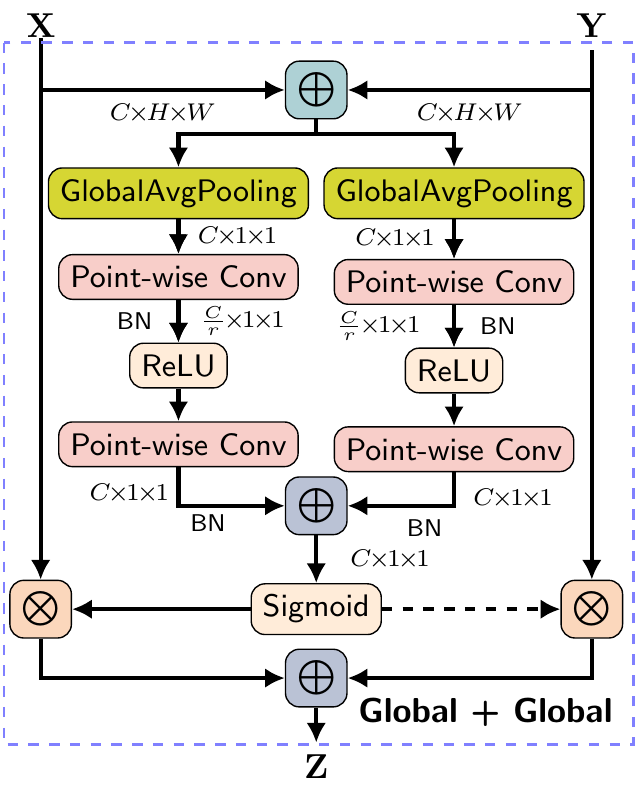}
  }
  \subfloat{
    \includegraphics[width=0.245\textwidth]{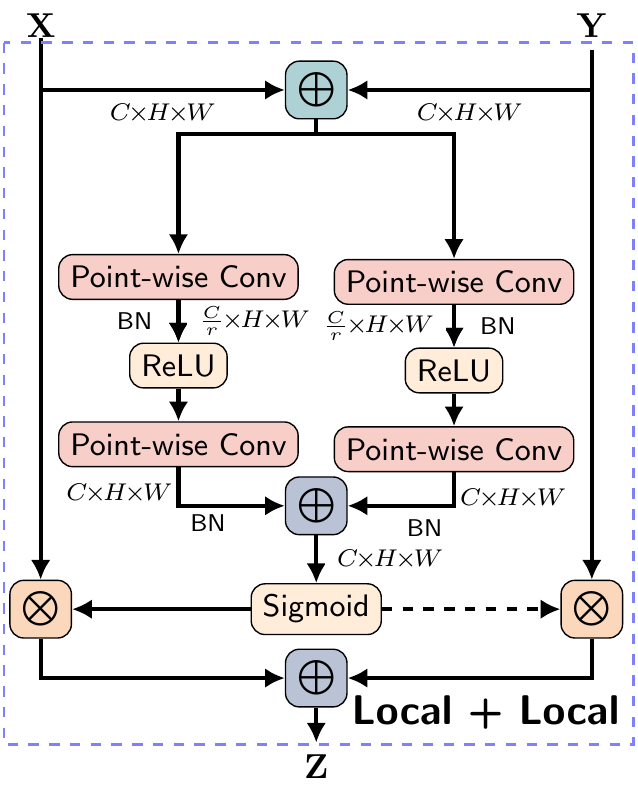}
  }\\[-.5\baselineskip]  
  \setlength{\belowcaptionskip}{-10pt}  
  \caption{Architectures for the ablation study on the impact of \textbf{contextual aggregation scale}.}
  \label{fig:ablationscale}
\end{figure}

To study the impact of multi-scale context aggregation, in \cref{fig:ablationscale}, we construct two ablation modules ``Global + Global'' and ``Local + Local'', in which the scales of the two contextual aggregation branches are set as the same, either global or local. 
The proposed AFF is dubbed as ``Global + Local'' here.
All of them have the same parameter number. The only difference is their \emph{context aggregation scale}.

Table~\ref{tab:agg_scale} presents their comparison on CIFAR-100, ImageNet, and StopSign on various host networks. 
It can be seen that the multi-scale contextual aggregation (Global + Local) outperforms single-scale ones in all settings. 
The results suggest that the multi-scale feature context is vital for the attentional feature fusion. 
\vspace*{-.5\baselineskip}

\setlength{\tabcolsep}{4pt}
\begin{table*}[htbp]
\begin{center}
\caption{Comparison of \textbf{contextual aggregation scales} in attentional feature fusion given the same parameter budget.
The results suggest that a mix of scales should always be preferred inside the channel attention module.}
\label{tab:agg_scale}
\vspace{-.5\baselineskip}
\small
\begin{tabular}{Sc Sc Sc Sc Sc Sc Sc Sc Sc Sc Sc Sc Sc Sc} 
\toprule  
\multirow{2}{*}{Aggregation Scale} & \multicolumn{4}{c}{InceptionNet on CIFAR-100} & \multicolumn{4}{c}{ResNet on CIFAR-100} & \multicolumn{4}{c}{ResNet + FPN on StopSign} & \multirow{2}{*}{\makecell{ResNet on\\ImageNet}} \\
\cmidrule(lr){2-5} \cmidrule(lr){6-9} \cmidrule(lr){10-13}
                & $b=1$          & $b=2$          & $b=3$          & $b=4$          & $b=1$          & $b=2$          & $b=3$          & $b=4$          & $b=1$ & $b=2$ & $b=3$ & $b=4$ & \\
\midrule 
Global + Global & 0.735          & 0.766          & 0.775          & 0.789          & 0.754          & 0.796          & 0.811          & 0.821          & 0.911 &  0.923     & 0.936      &  0.939 & 0.777\\
Local + Local   & 0.746          & 0.771          & 0.785          & 0.787          & 0.754          & 0.794          & 0.808          & 0.814          & 0.895 & 0.919      & 0.921      & 0.924 & 0.780\\
Global + Local  & \textbf{0.756} & \textbf{0.784} & \textbf{0.794} & \textbf{0.801} & \textbf{0.763} & \textbf{0.804} & \textbf{0.816} & \textbf{0.826} & \textbf{0.924} & \textbf{0.935} & \textbf{0.939} & \textbf{0.944} & \textbf{0.784}\\
\bottomrule
\end{tabular}
\end{center}
\vspace{-1\baselineskip}
\end{table*}

\setlength{\tabcolsep}{3pt}
\begin{table*}[!h]
\begin{center}
\caption{
Comparison of \textbf{context-aware level} and \textbf{feature integration strategy} in feature fusion \emph{given the same parameter budget}.
The results suggest that a fully context-aware and selective strategy should always be preferred for feature fusion. 
If no problem in optimization, we should adopt the iterative attentional feature fusion without hesitation for better performance.
}
\label{Tab:AblationFusion}
\vspace{-.5\baselineskip}
\small
\begin{tabular}{Sc Sc Sc Sc Sc Sc Sc Sc Sc Sc Sc Sc Sc Sc Sc} 
\toprule  
\multirow{2}{*}{Fusion Type} & \multirow{2}{*}{Context} & \multirow{2}{*}{Strategy} & \multicolumn{4}{c}{InceptionNet (Same Layer)} & \multicolumn{4}{c}{ResNet (Short Skip)} & \multicolumn{4}{c}{ResNet + FPN (Long Skip)} \\
\cmidrule(lr){4-7} \cmidrule(lr){8-11} \cmidrule(lr){12-15}
 & &              & $b=1$          & $b=2$          & $b=3$          & $b=4$          & $b=1$          & $b=2$          & $b=3$          & $b=4$ & $b=1$          & $b=2$          & $b=3$          & $b=4$ \\
\midrule 
Add & None &   \textbackslash        & 0.720          & 0.753          & 0.771          & 0.782         & 0.740          & 0.786          & 0.797          & 0.808  & 0.895 & 0.920 & 0.925 & 0.928 \\
Concatenation & None & \textbackslash & 0.725          & 0.749          & 0.772         & 0.779          & 0.742          & 0.782          & 0.793          & 0.798 & 0.897 & 0.909 & 0.925 & 0.939 \\
MS-GAU & Partially & Modulation
              & 0.751          & 0.774 & 0.788          & 0.795          & 0.766 & 0.803          & 0.815          & 0.819 & 0.917 & 0.926 & 0.937 & 0.941 \\

MS-SENet & Partially & Refinement
              & 0.752 & 0.780          & 0.790          & 0.798          & 0.765          & 0.799          & 0.814          & 0.820 & 0.915 & 0.929 & 0.940 & 0.940\\

MS-SA & Fully & Modulation
              & 0.756 & 0.779          & 0.790          & 0.798 & 0.761          & 0.801          & 0.814          & 0.822 & 0.920 & 0.932 & 0.938 & 0.941\\
AFF (\textbf{\textit{ours}}) & Fully & Selection
              & \textit{0.756}          & \textit{0.784}          & \textit{0.794} & \textit{0.801}          & \textit{0.763}          & \textit{0.804} & \textit{0.816} & \textbf{0.826} & \textit{0.924} & \textit{0.935} & \textit{0.939} & \textit{0.944}\\
iAFF (\textbf{\textit{ours}}) & Fully & Selection & \textbf{0.774} & \textbf{0.801} & \textbf{0.808} & \textbf{0.814} & \textbf{0.772} & \textbf{0.807} & \textbf{0.822} & / & \textbf{0.927} & \textbf{0.938} & \textbf{0.945} & \textbf{0.953}\\
\bottomrule
\end{tabular}
\end{center}
\vspace{-2\baselineskip}
\end{table*}

\subsubsection{Impact of Feature Integration Type}
Further, we investigate which feature fusion strategy is the best in \cref{Tab:RelatedSkip}.
For fairness, we re-implement these approaches based on the proposed MS-CAM for attention weights.
Since MS-CAM are different from their original attention modules, we add a prefix of "MS-" to these newly implemented schemes. 
To keep the parameter budget the same, here the channel reduction ratio $r$ in MS-GAU, MS-SE, MS-SA, and AFF is $2$, while $r$ in iAFF is $4$.
\vspace*{-.5\baselineskip}

\begin{figure}[htbp]
  \centering
  \captionsetup[subfloat]{farskip=0pt,captionskip=2pt}
  \subfloat[MS-GAU]{
    \includegraphics[height=0.18\textwidth]{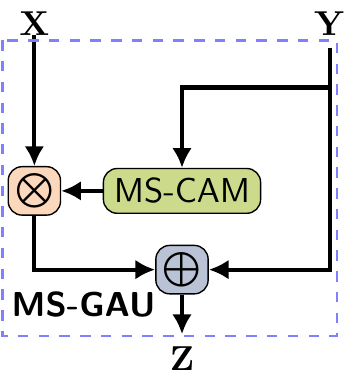}
  }\hspace*{-.6em}
  \subfloat[MS-SE]{
    \includegraphics[height=0.18\textwidth]{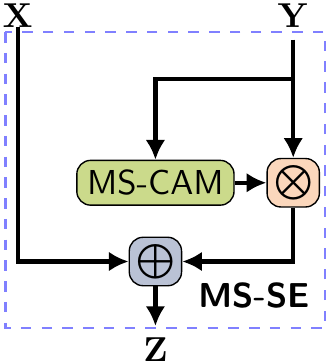}
  }\hspace*{-.4em}
  \subfloat[MS-SA]{
    \includegraphics[height=0.18\textwidth]{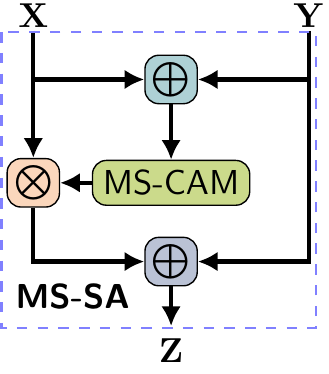} 
  }\\[-.5\baselineskip]
  \setlength{\belowcaptionskip}{-.5\baselineskip}  
  \caption{Architectures for ablation study on the impact of feature integration strategies}
\end{figure}

\cref{Tab:AblationFusion} provides the comparison results in three scenarios, from which it can be seen that:
1) compared to the linear approach, namely addition and concatenation, the non-linear fusion strategy with attention mechanism always offers better performance;
2) our fully context-aware and selective strategy is slightly but consistently better than the others, suggesting that it should be preferred for multiple feature integration;
3) the proposed iAFF approach is significantly better than the rest in most cases. 
The results strongly demonstrate our hypothesis that the early integration quality has a large impact on the attentional feature fusion, and another level of attentional feature fusion can further improve the performance.
However, this improvement may be obtained at the cost of increasing the difficulty in optimization.
We notice that when the network depth increases as $b$ changes from $3$ to $4$, the performance of iAFF-ResNet did not improve but degraded.

\subsubsection{Impact on Localization and Small Objects}

\begin{figure*}[htbp]
  \centering
  \subfloat{
    \includegraphics[width=.99\textwidth]{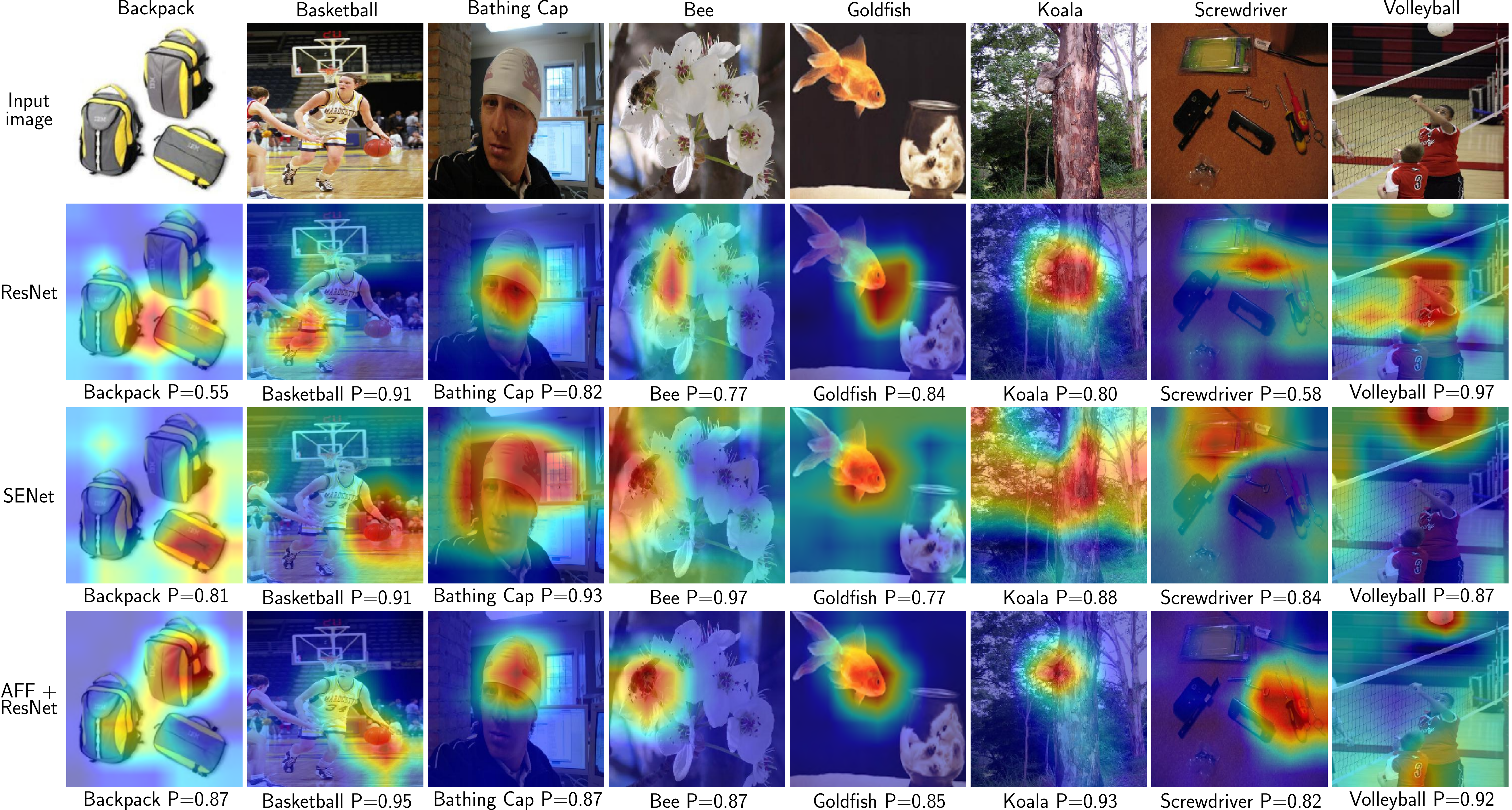}
  }\vspace*{-.5\baselineskip}

  \subfloat{
    \includegraphics[width=.99\textwidth]{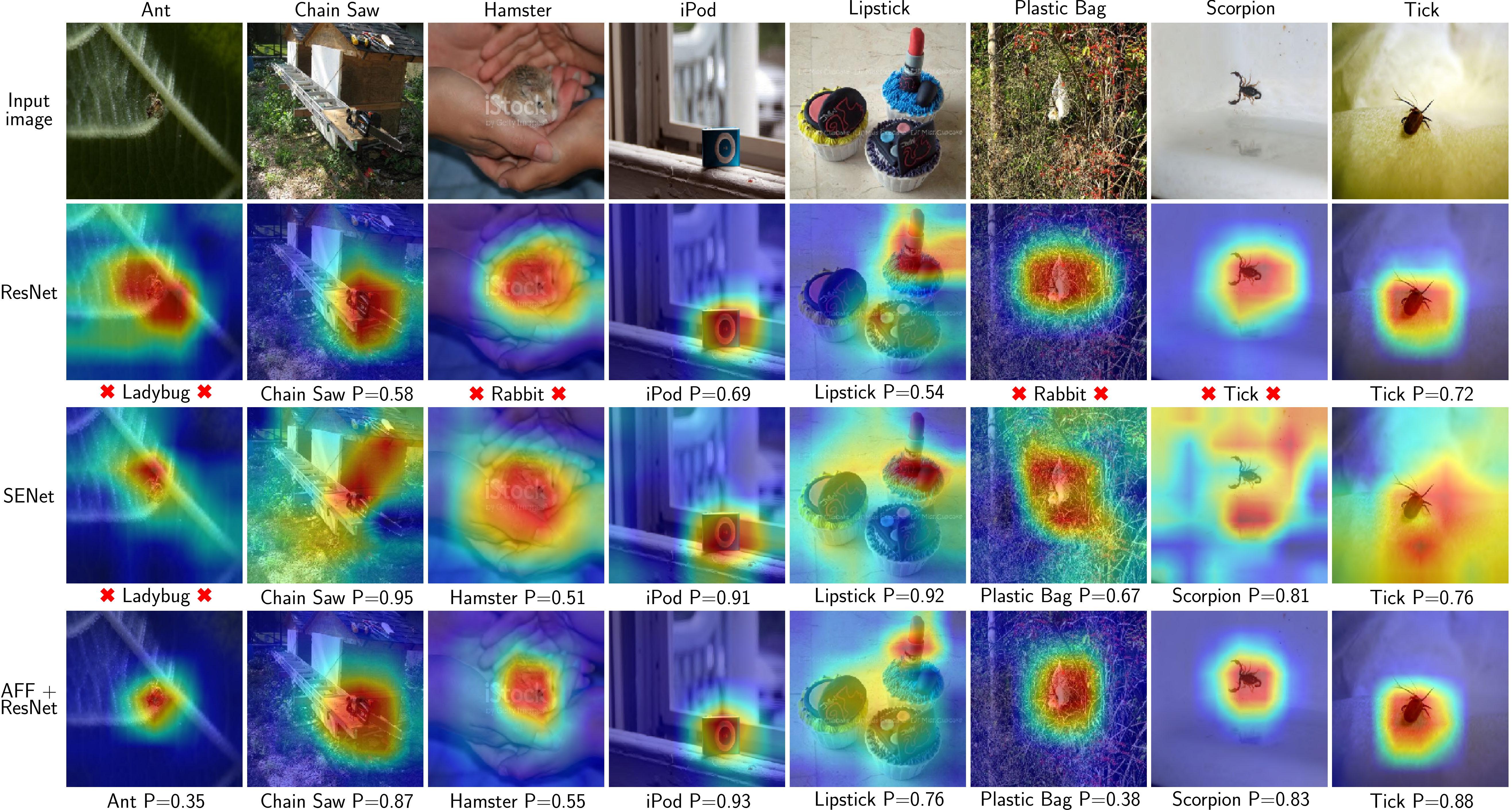}
  }  
  \setlength{\belowcaptionskip}{-10pt}  
  \caption{Network visualization with Grad-CAM. The comparison results suggest that the proposed MS-CAM is beneficial to the object localization and small object recognition.}
  \label{fig:visualization}
\end{figure*}

To study the impact of the proposed MS-CAM on object localization and small object recognition, we apply Grad-CAM \cite{IJCV20GradCAM} to ResNet-50, SENet-50, and AFF-ResNet-50 for the visualization results of images from the ImageNet dataset, which are illustrated in \cref{fig:visualization}.
Given a specific class, Grad-CAM results show the network's attended regions clearly. 
Here, we show the heatmaps of the predicted class, and the wrongly predicted image is denoted with the symbol \xmark.
The predicted class names and their softmax scores are also shown at the bottom of heatmaps.

From the upper part of \cref{fig:visualization}, it can be seen clearly that the attended regions of the AFF-ResNet-50 highly overlap with the labeled objects, which shows that it learns well to localize objects and exploit the features in object regions.
On the contrary, the localization capacity of the baseline ResNet-50 is relatively poor, misplacing the center of attended regions in many cases. 
Although SENet-50 are able to locate the true objects, the attended regions are over-large including many background components. 
It is because SENet-50 only utilizes the global channel attention, which is biased to the context of a global scale, whereas the proposed MS-CAM also aggregates the local channel context, which helps the network to attend the objects with fewer background clutters and is also beneficial to the small object recognition.
In the bottom half of \cref{fig:visualization}, we can clearly see that AFF-ResNet-50 can predict correctly on the small-scale objects, while ResNet-50 fails in most cases.

\subsection{Comparison with State-of-the-Art Networks}

\begin{figure*}[htbp]
  \centering
  \captionsetup[subfloat]{farskip=0pt,captionskip=0pt}  
  \subfloat[InceptionNet (Same layer)]{
    \includegraphics[height=0.25\textwidth]{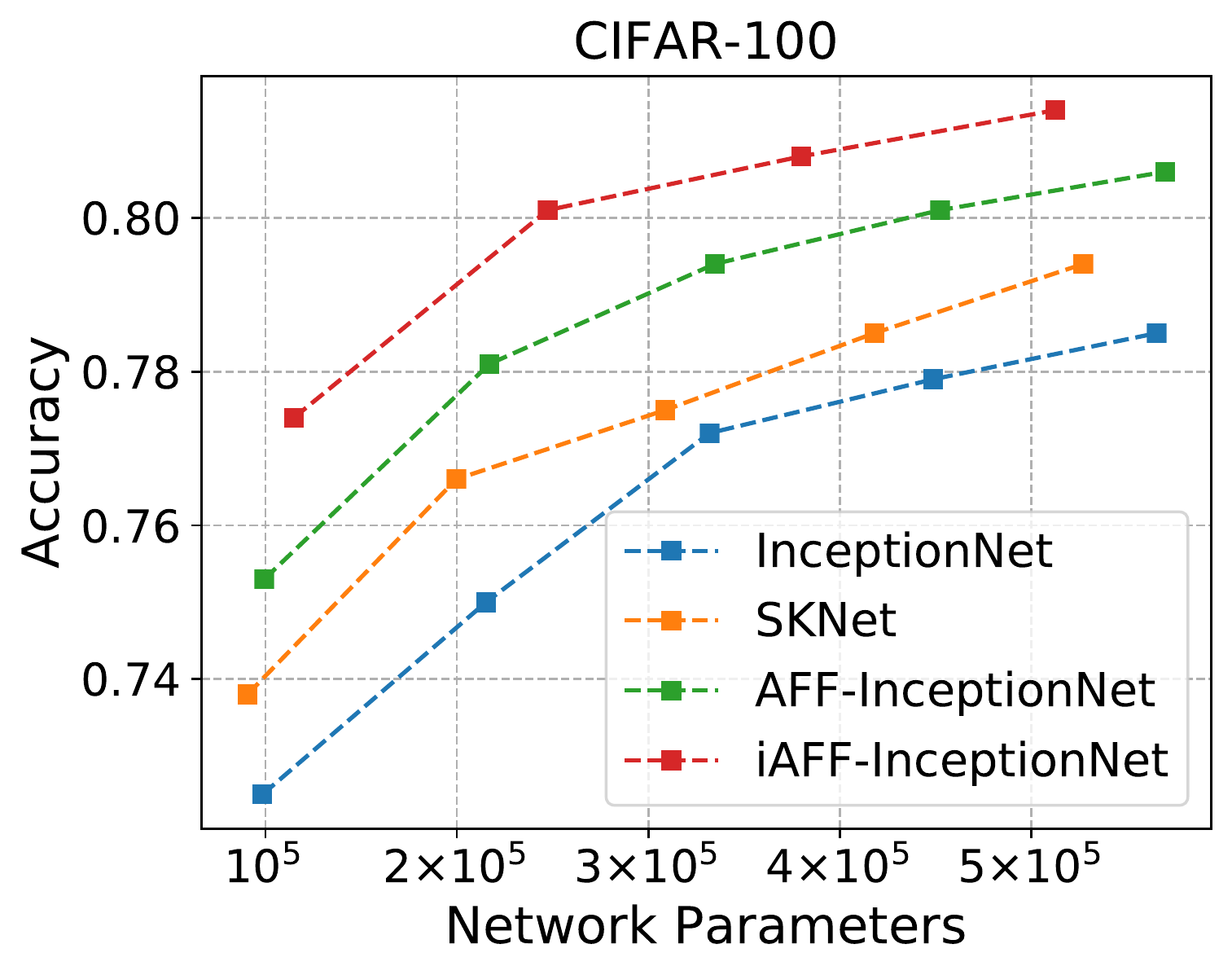}
  }
  \subfloat[ResNet (Short skip connection)]{
    \includegraphics[height=0.25\textwidth]{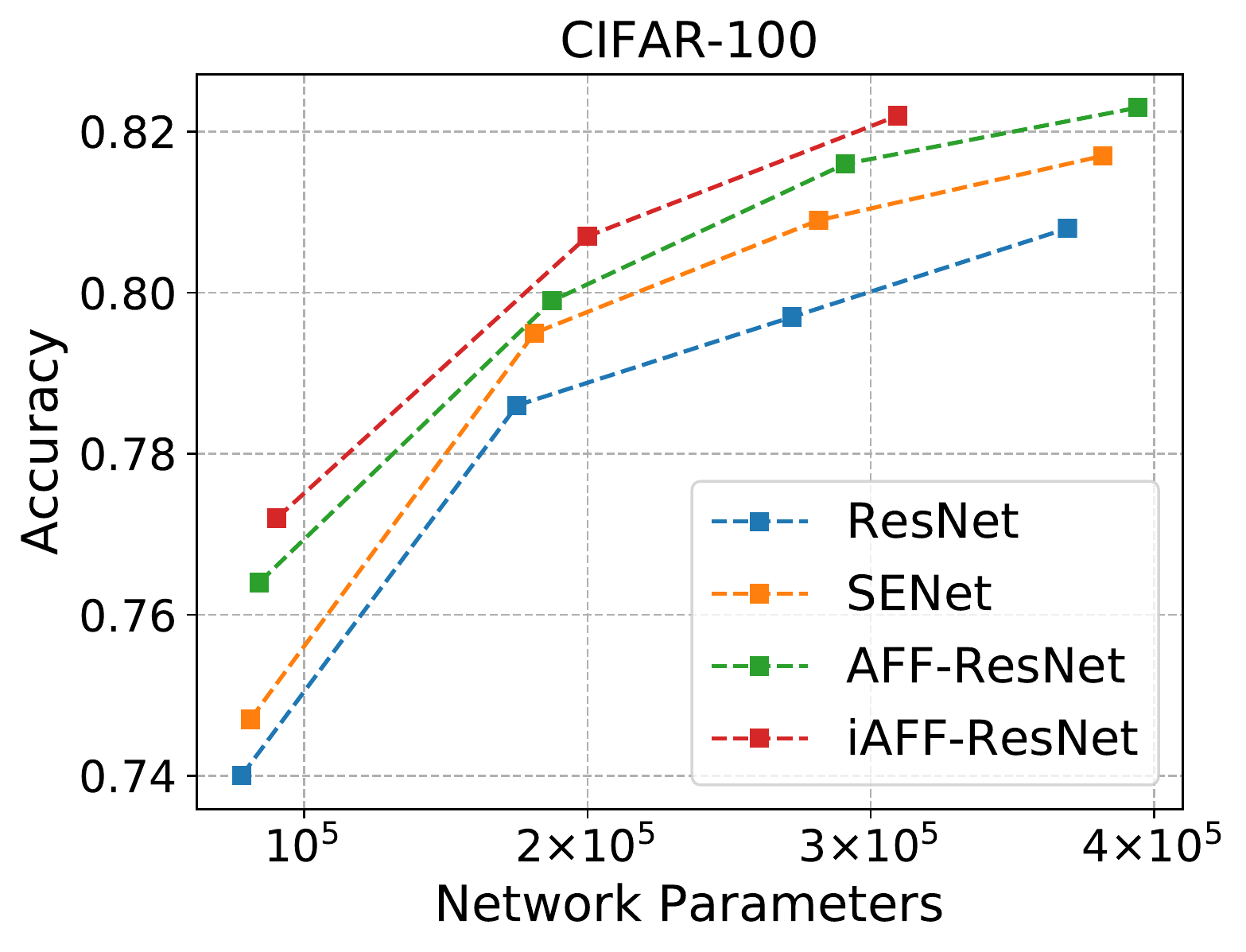}
    \label{subfig:sotashortskip}
  }
  \subfloat[FPN (Long skip connection)]{
    \includegraphics[height=0.25\textwidth]{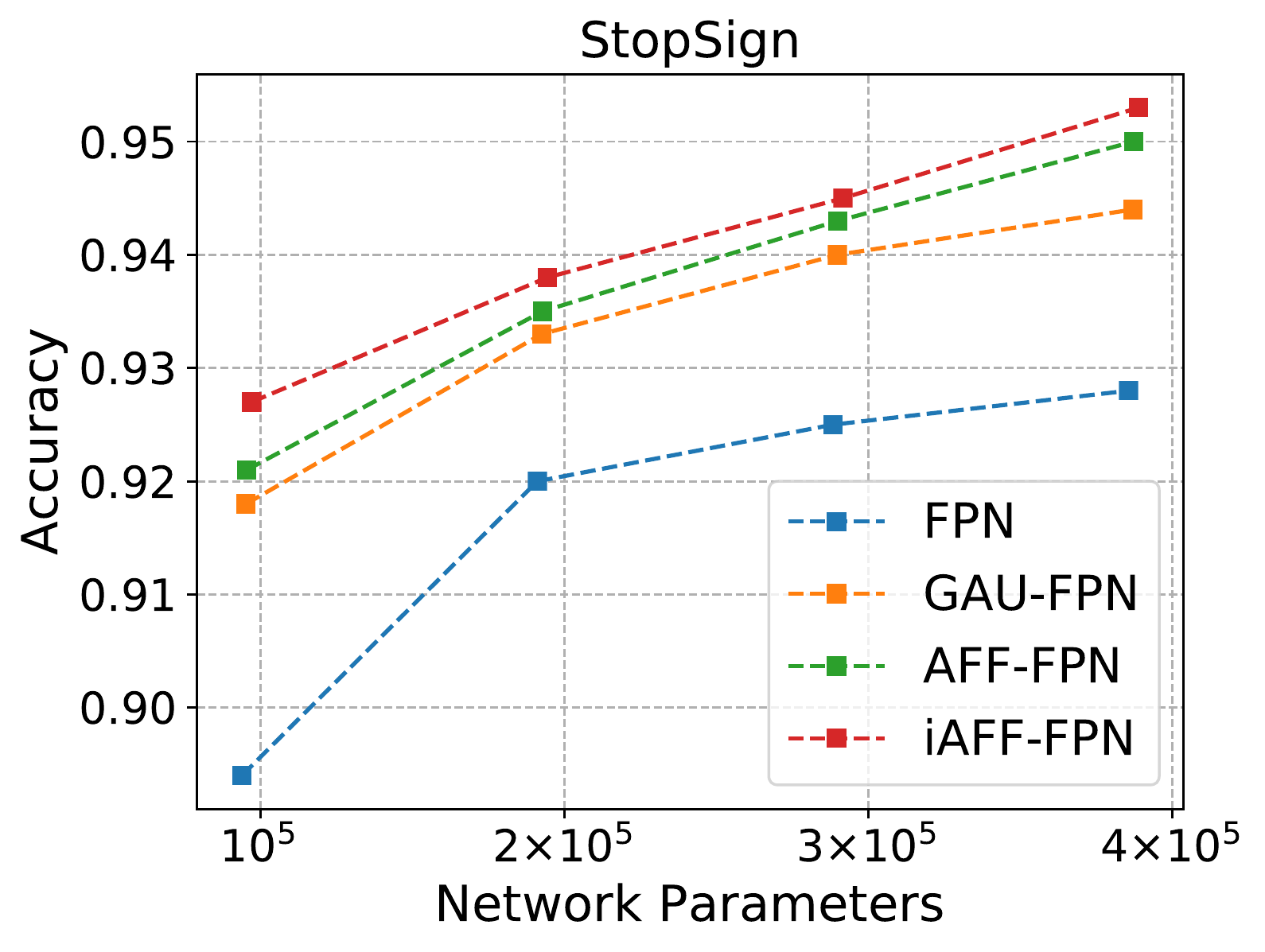}     
  }\\[-.5\baselineskip]
  \setlength{\belowcaptionskip}{-\baselineskip}  
  \caption{Compassion with baseline and other state-of-the-art networks with a gradual increase of network depth. }
  \label{fig:sota}
\end{figure*}

To show that the network performance can be improved by replacing original fusion operations with the proposed attentional feature fusion, we compare the AFF and iAFF modules with other attention modules based on the same host networks in different feature fusion scenarios. 
\cref{fig:sota} illustrates the comparison results with a gradual increase in network depth for all networks.
It can be seen that:
1) Comparing SKNet / SENet / GAU-FPN with AFF-InceptionNet / AFF-ResNet / AFF-FPN, we can see that our AFF or iAFF integrated networks are better in all scenarios, which shows that our (iterative) attentional feature fusion approach not only has superior performance, but a good generality.
We believe the improved performance comes from the proposed multi-scale channel contextual aggregation inside the attention module.
2) Comparing the performance of iAFF-based networks with AFF-based networks, it should be noted that the proposed iterative attentional feature fusion scheme can further improve the performance.
3) By replacing the simple addition or concatenation with the proposed AFF or iAFF module, we can get a more efficient network. 
For example, in Fig.~\subref*{subfig:sotashortskip}, iAFF-ResNet ($b=2$) achieves similar performance with the baseline ResNet ($b=4$), while only 54\% of the parameters were required.


Last, we validate the performance of AFF/iAFF based networks with state-of-the-art networks on ImageNet. 
The results are listed in \cref{tab:imagenet}.
The results show that the proposed AFF/iAFF based networks can improve performance over the state-of-the-art networks under much smaller parameter budgets.
Remarkably, on ImageNet, the proposed iAFF-ResNet-50 outperforms Gather-Excite-$\theta^{+}$-ResNet-101 \cite{NIPS18GENet} by 0.3\% with only 60\% parameters.
These results indicate that the feature fusion in short skip connections matters a lot for ResNet and ResNeXt. 
Instead of blindly increasing the depth of the network, we should pay more attention to the quality of feature fusion.


\setlength{\tabcolsep}{2pt}
\begin{table}[h]
\centering
\caption{Comparison on \textbf{ImageNet}}
\label{tab:imagenet}\vspace{-.5\baselineskip}
\small
\begin{tabular}{Sl Sc Sc} 
\toprule  
Architecture                                     & top-1 err.              & Params    \\
\midrule 
ResNet-101   \cite{CVPR16ResNetV1}                        & 23.2                   & 42.5 M                   \\
Efficient-Channel-Attention-Net-101 \cite{CVPR20ECANet}       & 21.4                   & 42.5 M                   \\
Attention-Augmented-ResNet-101 \cite{ICCV19AugmentedConv} & 21.3                   & 45.4 M                    \\
SENet-101 \cite{CVPR18SENet}                              & 20.9                   & 49.4 M                    \\
Gather-Excite-$\theta^{+}$-ResNet-101 \cite{NIPS18GENet}  & 20.7                   & 58.4 M                   \\
Local-Importance-Pooling-ResNet-101 \cite{ICCV19LIP}                           & 20.7                   & 42.9 M                    \\
\textbf{\textit{AFF-ResNet-50 (ours)}}                   & \textit{\textbf{20.9}} & \textit{\textbf{30.3 M}} \\
\textbf{\textit{AFF-ResNeXt-50-32x4d (ours)}}                   & \textit{\textbf{20.8}} & \textit{\textbf{29.9 M}} \\
\textbf{\textit{iAFF-ResNet-50 (ours)}}                   & \textit{\textbf{20.4}} & \textit{\textbf{35.1 M}} \\
\textbf{\textit{iAFF-ResNeXt-50-32x4d (ours)}}                   & \textit{\textbf{20.2}} & \textit{\textbf{34.7 M}} \\
\bottomrule
\end{tabular}
\vspace*{-\baselineskip}
\end{table}


\section{Conclusion}

We generalize the concept of attention mechanisms as a selective and dynamic type of feature fusion to most scenarios, namely the same layer, short skip, and long skip connections as well as information integration inside the attention mechanism.
To overcome the semantic and scale inconsistency issue among input features, we propose the multi-scale channel attention module, which adds local channel contexts to the global channel-wise statistics.  
Further, we point out that the initial integration of received features is a bottleneck in attention-based feature fusion, and it can be alleviated by adding another level of attention that we call iterative attentional feature fusion.
We conducted detailed ablation studies to empirically verify the individual impact of the context-aware level, the feature integration type, and the contextual aggregation scales of our proposed attention mechanism.
Experimental results on both the CIFAR-100 and the ImageNet dataset show that our models outperform state-of-the-art networks with fewer layers or parameters per network, which suggests that one should pay attention to the feature fusion in deep neural networks and that more sophisticated attention mechanisms for feature fusion hold the potential to consistently yield better results.

\section*{Acknowledgement}


The authors would like to thank the editor and anonymous reviewers for their helpful comments and suggestions, and also thank @takedarts on Github for pointing out the bug in our CIFAR-100 code.
This work was supported in part by the National Natural Science Foundation of China under Grant No. 61573183, the Open Project Program of the National Laboratory of Pattern Recognition (NLPR) under Grant No. 201900029, the Nanjing University of Aeronautics and Astronautics PhD short-term visiting scholar project under Grant No. 180104DF03, the Excellent Chinese and Foreign Youth Exchange Program, China Association for Science and Technology, China Scholarship Council under Grant No. 201806830039.

\newpage
\clearpage
{\small
\bibliographystyle{ieee_fullname}
\bibliography{reference}
}


\section*{Appendix}

\subsection*{Implementation Details}

All network architectures in this work are implemented based on MXNet~\cite{NIPSW15MXNet} and GluonCV \cite{CVPR19GluonCV}. 
Since most of the experimental architectures cannot take advantage of pre-trained weights, each implementation is trained from scratch for fairness.
We have introduced most of the experimental settings in Table 2 of the manuscript. 
Here, in the supplemental document, we introduce the left settings that not mentioned before. 

For the experiments on the CIFAR-100 dataset, the weight decay is 1e-4, and we decay the learning rate by a factor of 0.1 at epoch 300 and 350.

For the experiments on the ImageNet, we use the label smoothing trick and a cosine annealing schedule for the learning rate without weight decay.

\begin{figure*}[htbp]
  \centering
\includegraphics[width=0.99\textwidth]{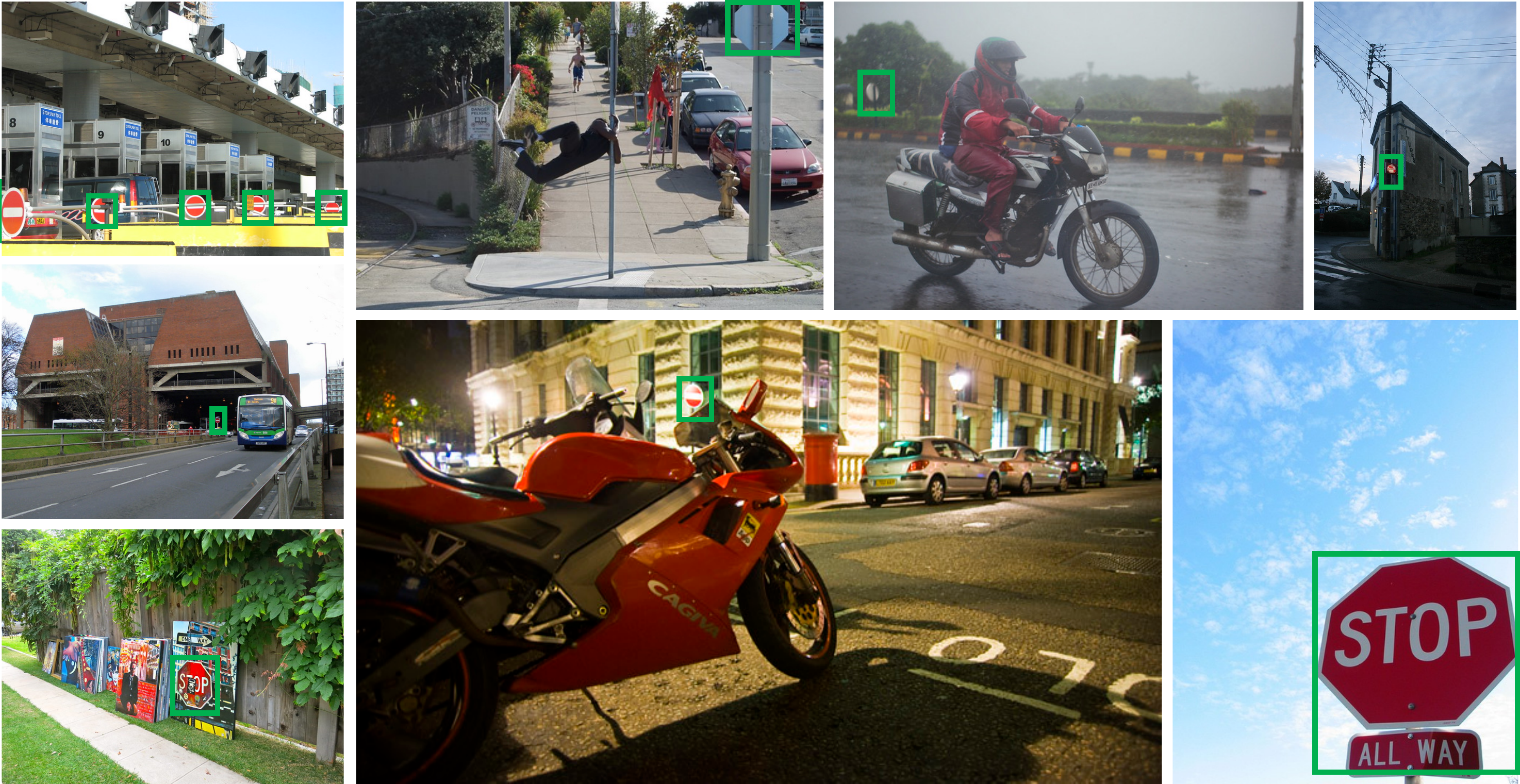}
  \caption{Illustration for the StopSign dataset}
  \label{fig:stopsign}
\end{figure*}

For the semantic segmentation experiment, the StopSign dataset is a subset of the COCO dataset \cite{ECCV14COCO}, which has a large scale variation issue, as shown in \cref{fig:stopsign}. 
We use the cross entropy as loss function and the mean intersection over union (mIoU) as evaluation metric.

It should be noted that the proposed networks in Table 5 and Table 6 are trained with mixup \cite{ICLR18mixup}.
The rest experiments, including all the ablation studies and the experimental results in Figure 7 (in the manuscript) are trained without mixup.

\subsection*{Local and Global Fusion Strategies}

\begin{figure*}[htbp]
  \centering
  \captionsetup[subfloat]{farskip=0pt,captionskip=0pt}  
  \subfloat[Half-AFF]{
    \includegraphics[width=0.375\textwidth]{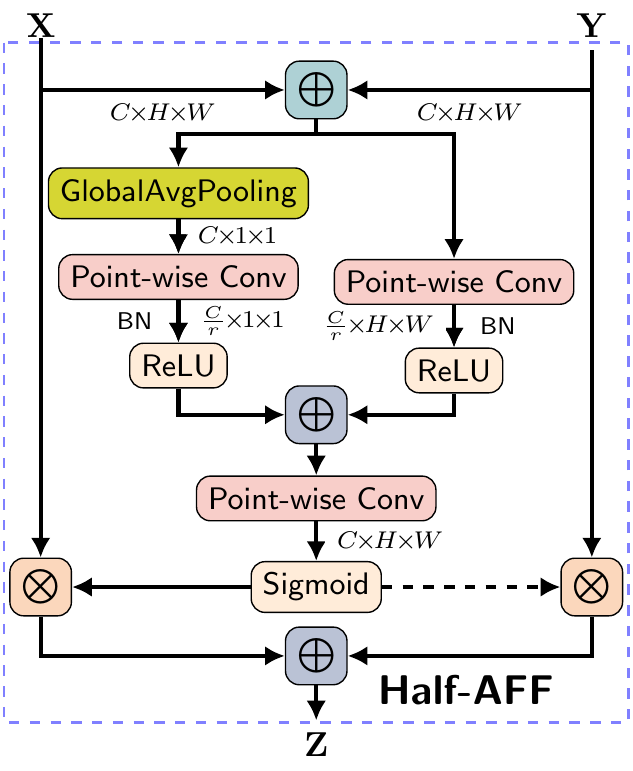}
  }\hfil
  \subfloat[Concat-AFF]{
    \includegraphics[width=0.375\textwidth]{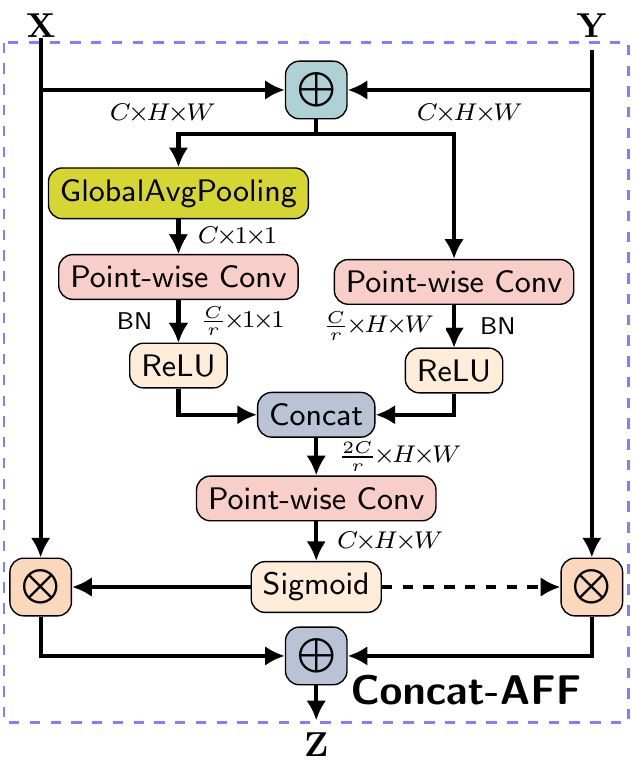}
  }

  \subfloat[AFF]{
    \includegraphics[width=0.425\textwidth]{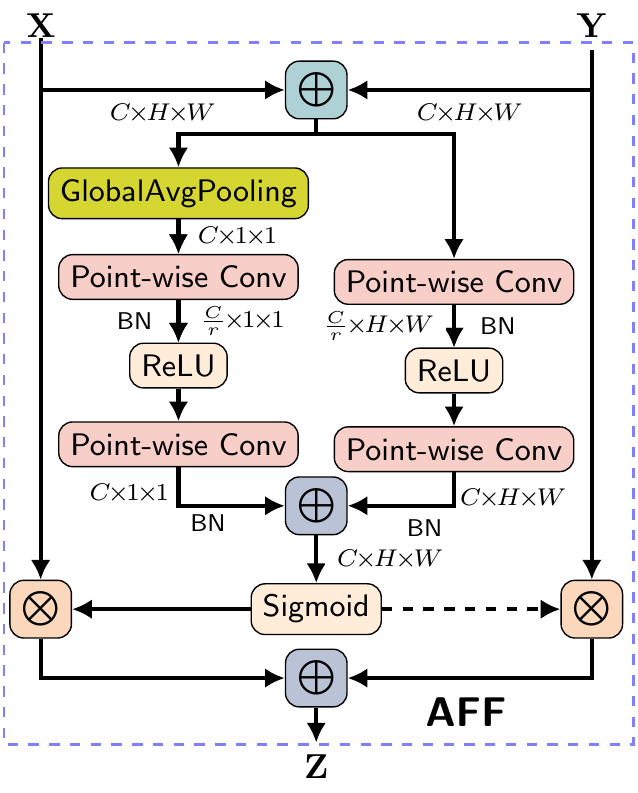}    
    \label{subfig:coupledaff} 
  }\hfil
  \subfloat[Recursive AFF]{
    \includegraphics[width=0.425\textwidth]{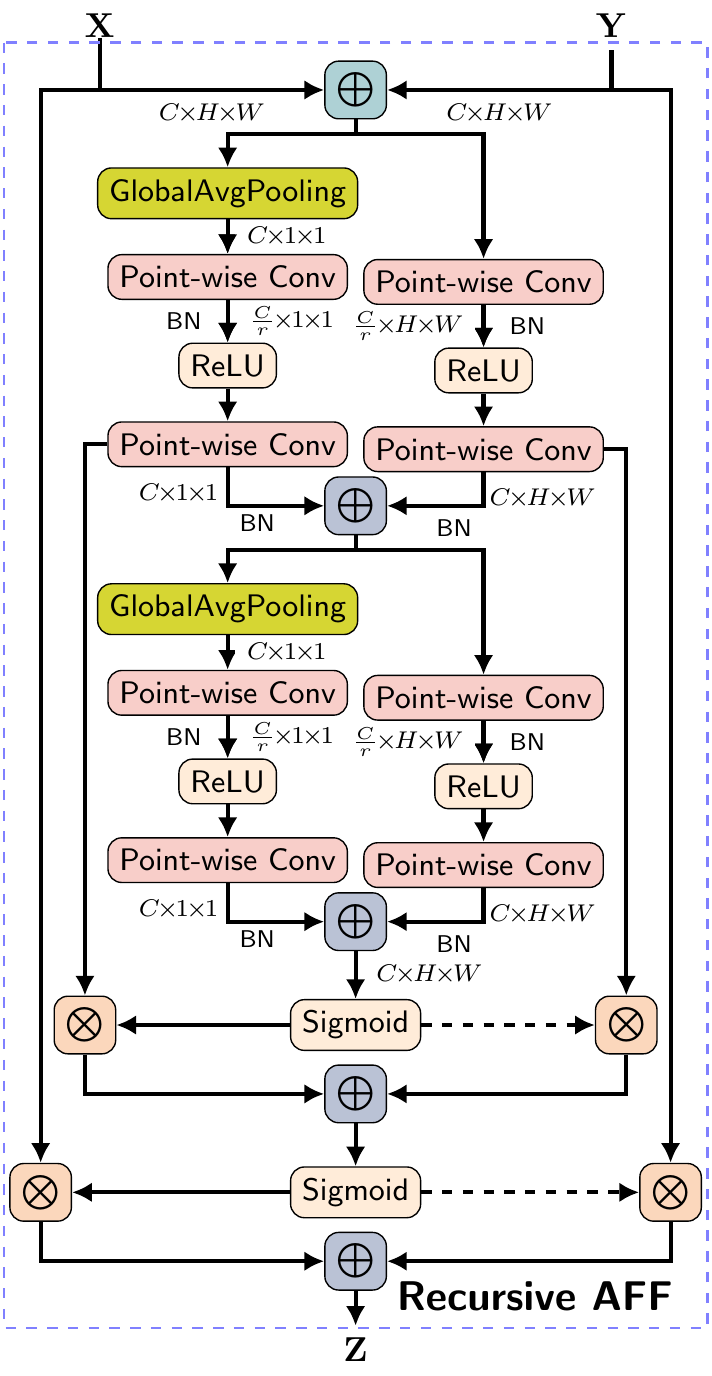} 
    \label{subfig:rAFF}    
  }
  \caption{Architectures for the ablation study on the fusion manner of the local and global channel contexts.}
  \label{fig:ablationcontextfuse}
\end{figure*}

We also investigate the fusion strategy for the local and global contexts inside the attention module. 
We explored four strategies as shown in \cref{fig:ablationcontextfuse}, in which:
\begin{enumerate}
  \item Half-AFF, AFF, and Iterative AFF apply addition to fuse the local and global contexts, which allocate the same weights (a constant 0.5) for local and global contexts. 

  \item Concat-AFF concatenates the local and global contexts followed by a point-wise convolution, in which the fusing weights are learned during training and fixed after training. 

  \item Recursive AFF allocates dynamic fusion weights for the local and global contexts during inference based on the proposed MS-CAM. 
\end{enumerate}

\cref{tab:contextfusion} provides the experimental results of these modules on CIFAR-100, from which it can be seen that the iterative AFF (iAFF) module presented in the manuscript achieves the best performance. 
On the contrary, the Recursive AFF which can dynamically allocate fusion weights for local and global contexts are almost the worst among these modules. 
We believe the reason is that Recursive AFF has two successive nested Sigmoid functions (see Fig.~\subref*{subfig:rAFF}), which increases the difficulty in optimization due to Sigmoid's saturation function form, whereas the iterative AFF presented in the manuscript does not suffer from this problem.

AFF and Concat-AFF have a very similar performance. 
Therefore, for simplicity, we choose the squeeze-and-excitation form (current MS-CAM module) instead of the Inception-style form (Concat-AFF) for the proposed attentional feature fusion. 
In future work, we will investigate their performance difference on larger datasets like ImageNet. 
However, this point is not the main issue that we would like to discuss in the manuscript, so we didn't include this part in the manuscript.

\begin{table*}[htbp]
\caption{Results for the ablation study on the fusion manner of the local and global channel contexts on CIFAR-100}
\label{tab:contextfusion}
\vspace*{.5\baselineskip}
\centering
\begin{tabular}{c c c c c} 
\toprule 
\multirow{2}{*}{Module}        & \multirow{2}{*}{\makecell{Fusion weights of local and\\global channel contexts}} & \multirow{2}{*}{$b = 1$}        & \multirow{2}{*}{$b = 2$}        & \multirow{2}{*}{$b = 3$} \\
& & & & \\
\midrule 

Half-AFF      & Constant, 0.5 for each & 0.759          & 0.798          & 0.813 \\
Concat-AFF    & Learned, fixed after training & 0.765          & 0.792          & 0.817 \\
AFF           & Constant, 0.5 for each & 0.764          & 0.799          & 0.816 \\
\midrule 
Recursive AFF & \makecell{Dynamic, depending on the local\\and global channel contexts} & 0.764          & 0.797          & 0.812 \\
Iterative AFF & Constant, 0.5 for each & \textbf{0.772} & \textbf{0.807} & \textbf{0.822} \\
\bottomrule
\end{tabular}
\end{table*}

\subsection*{Analysis of FLOPs}

The point-wise convolution inside our multi-scale channel attention module can bring additional FLOPs, but at a marginal level, not a significant magnitude. 
The FLOPs of our AAF-ResNet-50 is 4.3 GFlops, and the Flops of ResNet-50 in our implementation is 4.1 GFlops. 
Actually, depending on how many tricks are used in ResNet, the Flops of ResNet-50 can vary from 3.9 GFlops to 4.3 GFlops \cite{CVPR19GluonCV}. 
Therefore, taking ResNet-50 vs our AFF-ResNet-50 for example, integrating the AFF module only brings additional 4.88\% Flops from 4.1 GFlops to 4.3 GFlops. 
Considering the performance boost by the AFF module, we think additional 4.88\% Flops is a good trade-off.

Given an output channel number $C$ and the size $H \times W$ of a output feature map, if the input channel number and output channel number are the same, the Flops of a $3 \times 3$ convolution layer is $18C^2HW$ (multiplication and addition), and a ResBlock consists of two or three convolution layers. 
Meanwhile, the Flops of two point-wise convolutions of a bottleneck structure is $\frac{2}{r}C^2HW$, where $r = 4$ or $r = 16$ depending on the dataset and network.
Therefore, comparing the Flops of convolutions in the host network, the Flops brought by the AFF module is marginal.

In \cref{tab:flops}, we list the Flops of convolutions in BasicResBlock / BottleneckResBlock, Flops of point-wise convolution in our AFF module, and the relative increasing percentage. 
It can be seen that the maximum additional flops brought by the AFF module in percentage is around 7.7\% if we use AFF module in each ResBlock from beginning to end. 
However, it is not necessary to replace every ResBlock with AFF-ResBlock. 
In our AFF-ResNet, we do this replacement from the middle of the network (last two stages), while leaving the first two stages of the original BottleneckResBlock. It further reduces the Flops of AFF-ResNet-50.

\setlength{\tabcolsep}{3pt}
\begin{table*}[htbp]
\centering
\caption{Additional Flops brought by the proposed AFF module in an AFF-ResBlock
}
\label{tab:flops}
\small
\begin{tabular}{Sc Sc Sc Sc Sc} 
\toprule  
\multirow{2}{*}{\makecell{ResBlock Type}} & \multirow{2}{*}{\makecell{Layer doubling \\ channel number ?}} & \multirow{2}{*}{\makecell{Flops of Conv\\in ResBlock}} & \multirow{2}{*}{\makecell{Flops of Point-wise\\Convin AFF module}} & \multirow{2}{*}{\makecell{Percentage}}\\
& & & & \\
\midrule 
\multirow{2}{*}{\makecell{BasicResBlock\\(CIFAR, $r = 4$)}} & Yes & $27C^2HW$ & $C^2HW$ & 3.70\%\\
& No & $36C^2HW$ & $C^2HW$ & 2.78\% \\
\multirow{2}{*}{\makecell{BottleneckResBlock\\(ImageNet, $r = 16$)}} & Yes & $51C^2HW$ & $4C^2HW$ & 7.84\% \\
& No & $52C^2HW$ & $4C^2HW$ & 7.69\% \\
\bottomrule
\end{tabular}
\end{table*}

To conclude, the AFF module will bring additional Flops but at a marginal level, around 3\% to 5\%. 
We think it is a good trade-off since the AFF module boosts the representation power of the convolution networks.


\end{document}